\documentclass{article}
\usepackage[T1]{fontenc}

\usepackage{graphicx}
\usepackage{url}

\usepackage{authblk}

\usepackage[inkscapeformat=png]{svg}
\usepackage{color}

\usepackage{fancyhdr}

\fancypagestyle{plain}{%
   \fancyhf{} 
\fancyfoot[L]{\footnotesize Preprint submitted to  the \textit{17th International Conference on Metadata and Semantics Research 2023} and published as full, revised article in \textit{Garoufallou, E., Sartori, F. (eds) Metadata and Semantic Research. MTSR 2023. Communications in Computer and Information Science, vol 2048.} DOI: \url{https://doi.org/10.1007/978-3-031-65990-4_14}}

}

\date{}

\begin{document}

\title{Ontologies for Models and Algorithms in Applied Mathematics and Related Disciplines
}

\author[1]{Björn Schembera\thanks{Corresponding author: \textit{bjoern.schembera@ians.uni-stuttgart.de}}}
\author[2]{Frank Wübbeling}
\author[2]{Hendrik Kleikamp}
\author[3]{Christine Biedinger}
\author[3]{Jochen Fiedler}
\author[4]{Marco Reidelbach}
\author[5]{Aurela Shehu}
\author[5]{Burkhard Schmidt}
\author[5]{Thomas Koprucki}
\author[6]{Dorothea Iglezakis}
\author[1,7]{Dominik Göddeke}

\affil[1]{\small{Institute of Applied Analysis and Numerical Simulation, University of Stuttgart}}
\affil[2]{\small{Institute of Applied Mathematics: Analysis and Numerics, University of Münster}}
\affil[3]{\small{Fraunhofer Institute for Industrial Mathematics, Kaiserslautern}}
\affil[4]{\small{Mathematics of Complex Systems, Zuse Institute Berlin}}
\affil[5]{\small{Weierstrass Institute for Applied Analysis and Stochastics, Berlin}}
\affil[6]{\small{University Library, University of Stuttgart}}
\affil[7]{\small{Stuttgart Center for Simulation Science (SC SimTech), University of Stuttgart}}

\maketitle              
\begin{abstract}
In applied mathematics and related disciplines, the modeling-simulation-optimization workflow is a prominent scheme, with mathematical models and  numerical algorithms playing a crucial role. For these types of mathematical research data, the Mathematical Research Data Initiative has developed, merged and implemented ontologies and knowledge graphs. This contributes to making mathematical research data FAIR by introducing semantic technology and documenting the mathematical foundations accordingly.
Using the concrete example of microfracture analysis of porous media, it is shown how the knowledge of the underlying mathematical model and the corresponding numerical algorithms for its solution can be represented by the ontologies.

\end{abstract}
\section{Introduction}
\label{sec:Introduction}

Mathematical research data have a multitude of appearances.
Classically, these are documents with mathematical proofs and formulae,
but increasingly models, algorithms and software as well as the associated generated data 
(numerical, symbolic) are included~\cite{MaRDI2022,Boege2022}.
Those data stem from the mathematical core sciences as well as from 
the applied sciences, e.g., engineering or physics. 
\begin{figure}[h!]

    \includegraphics[width=\textwidth]{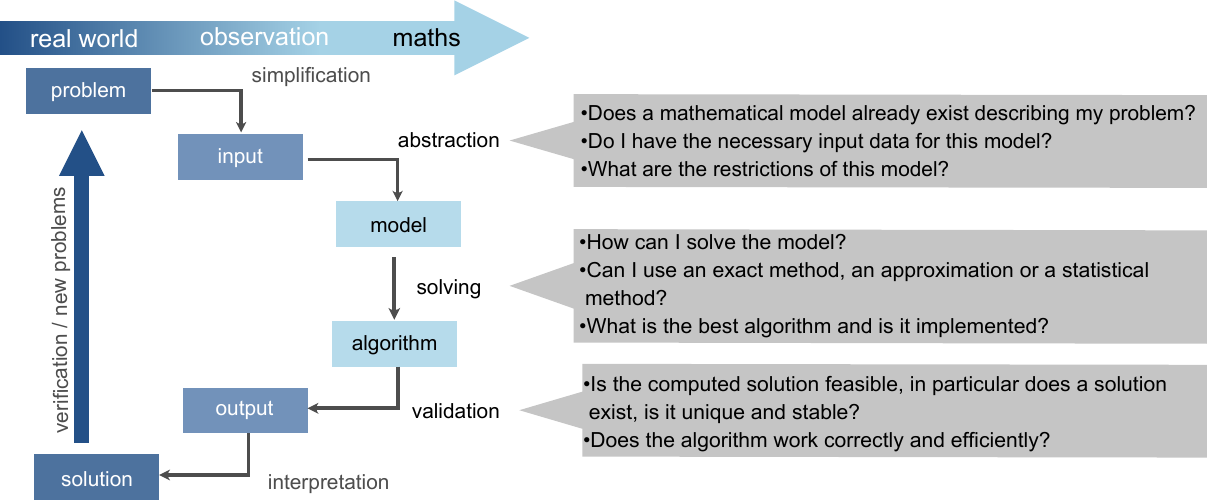}

   \caption{Typical MSO workflow  and resulting competency questions~\cite{MaRDI2022}.}\label{fig:MSO_Workflow}
\end{figure}

Consider the following situation as a  guiding example: We want to examine a sample of some porous medium, like open-pored asphalt concrete, and analyze it using a micro \textit{X-ray computed tomography} (X-RCT) scan to detect micro fractures in the material~\cite{Ruf2020}. 

The measurement process can be mathematically modeled via a Radon transform in the following sense: When an X-ray travels on a line through an object it will be attenuated by the material on this line. This attenuation depends on the density of the material, which we want to reconstruct. Mathematically, the signal measured in the detector can now be expressed as a Radon transform, the so called X-ray transform of the density function. 

Hence, to reconstruct the fracture images, an algorithm for inversion of the X-ray transform has to be applied to the observed data.
Among others, the choice of the algorithm depends on the measured data and the properties of the model, like the coordinate system used. These metadata are often not stored systematically, causing violations of the FAIR principles~\cite{Wilkinson2016}, as the reusability cannot be guaranteed.

Consequently, researchers who are interested in applying X-RCT, possibly in other research areas like archaeology or biomedicine, cannot simply reuse, but may have to redo the literature search on algorithms, software implementations and parameters from scratch again.
Due to their origin from engineering,
data from different domains are not linked to the underlying general mathematical concept. Hence, synergies between applications are not exploited, although the basic mathematical model may be exactly the same. 
The creation of a \textit{Knowledge Graph} (KG) including models, algorithms, related literature and further metadata is in the scope of this paper.

In general, questions arising in a typical \textit{Modeling-Simulation-Optimization} (MSO) workflow like shown in fig.~\ref{fig:MSO_Workflow} should be picked up. These include the existence of models, availability of solution algorithms, input or observation data or model validity. Usually, answering those questions requires a huge amount of effort that can be reduced if the required information is accessible and discoverable from a single starting point in a KG, shareable by a unique identifier, and browseable through a portal.

In the following sections we present our work on 
the ontologies for mathematical models (MathModDB) and for algorithms (AlgoData) and how they are connected, as the basis for a KG. The work presented is conducted within the \textit{Mathematical Research Data Initiative} (MaRDI) project of the \textit{German National Research Data Infrastructure} (NFDI). The NFDI is a part of a nation-wide coordinated effort that aims at setting up data infrastructure and semantic technologies to foster the FAIR principles.
MaRDI is the NFDI consortium for mathematics with the mission to develop a robust research data infrastructure for mathematics and bridging towards other scientific disciplines~\cite{MaRDI2022}. 
\section{Related Work}
Using ontologies for the semantic representation of mathematical knowledge is not yet widespread. Among the few examples are \textit{Open Mathematical Documents}~(OMDoc)~\cite{Kohlhase:OMDoc}, which is a semantic markup format and ontology for mathematical documents consisting of classes such as \textit{Definition}, \textit{Theorem} and \textit{Proof}. Furthermore, the ontology of mathematical knowledge concepts (OntoMath) was initially developed to unify terminology in the field of mathematics and introduces a taxonomic approach for mathematical knowledge representation~\cite{Elizarov2017}. It was developed for educational purposes and eventually merged into the OntoMathEdu ontology, which can be used to model "prerequisite" relations to create educational plans of mathematics~\cite{Falileeva2020}. 

Ontologies for models and algorithms remain restricted to the implementation aspects of computational models in plasma physics~\cite{Snytnikov2020} or mathematical models in biological processes~\cite{Inizan2021}. 
A more general way to represent the semantics in mathematical models are \textit{Model Pathway Diagrams} (MPDs). They are based on \textit{Quantities} (represented by terms) which are connected by \textit{Laws} (represented by equations)~\cite{Koprucki2018}. The work presented in this paper can be seen as a generalization of MPDs with respect to more detailed information on mathematical models.

In \textit{Wikidata}~\cite{Vrandevic2014}, mathematical models and algorithms are semantically described using classes like \textit{computational problem}, \textit{algorithm} or \textit{mathematical model} and properties such as \textit{defining formula} or \textit{computes solution to}. However, until now a thorough standardization is missing. For example, the \textit{filtered back projection} algorithm (wikidata:Q20665529) is described as an instance of \textit{algorithm} and subclass of \textit{Radon transfer}, but no information about the \textit{defining formula} and other important aspects is provided. 

Within the NFDI, ontologies and KGs play a major role as semantic technology to implement FAIR principles~\cite{Koepler2021}. Thus, in many NFDI consortia  there are efforts to use these to describe and connect research assets. 
For our work, \textit{NFDI4Ing}~\cite{Schmitt2020} is particularly relevant. The ontology \textit{metadata4ing}~\cite{metadata4ing} (m4i) developed by that project describes scientific processes with a focus on the engineering domain. Conceptually, m4i is based on the metadata scheme EngMeta~\cite{Schembera2018,Schembera2020}, on existing terminologies~\cite{Missier2013,QUDT2023,Horsch2020} and on top-level ontologies~\cite{Otte2022,Guha2016}. 

In the following, two ontologies that we develop for mathematical models and numerical algorithms are presented, that aim to express as much semantic information as reasonable, but at the same time being compact and simple.

\section{A Combined Ontology for Models and Algorithms}
\label{sec:KnowledgeGraphs}
\begin{figure}[h]

    \includegraphics[width=\textwidth]{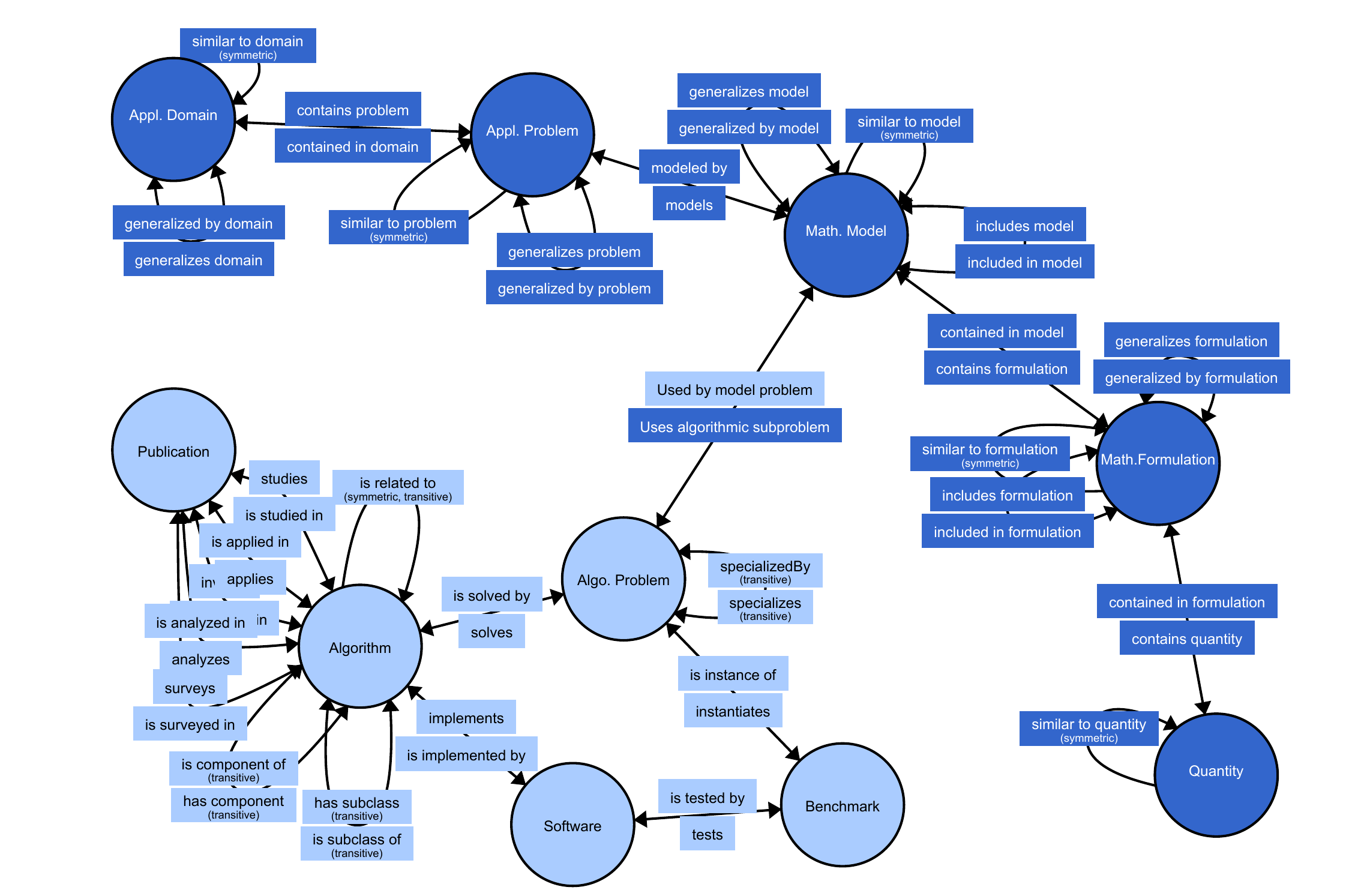}  
    \caption{The ontologies for algorithms (AlgoData, light blue) and models (MathModDB, dark blue). The ontologies are connected between the \textit{Mathematical Model} and the \textit{Algorithmic Problem} classes.}\label{fig:Ontology}
\end{figure}

In this section, we describe the evolution and development of ontologies for models and numerical algorithms (a preliminary version has been published in~\cite{Schembera2023_CoRDI}). Furthermore, we explain how these two graphs are connected (cf. fig.~\ref{fig:Ontology}).

\paragraph{\textbf{Models}} are fundamental mathematical research data (cf. sec.~\ref{sec:Introduction} and  \cite{KoTa2016LNCS}), allowing to abstract, formalize, analyze and ultimately understand complex phenomena from nature and technology. As depicted in fig.~\ref{fig:MSO_Workflow}, models are -- together with the algorithms -- at the heart of the typical MSO workflow. They lie at the intersection of mathematics and its fields of application.  Models from different application areas often share the same mathematical characteristics, and in some cases even structurally identical models are used, differing only in the quantities involved or the problems they are modeling \cite{KoTa2016LNCS}. This universality of mathematical models justifies the construction of the \textit{Mathematical Models Ontology} (MathModDB) to improve the semantic representation of knowledge in mathematics.  

By analyzing use cases and their workflows, such as, amongst others, X-RCT  (cf. sec.~\ref{sec:Introduction}), it is apparent that the following classes are relevant for a comprehensive description of mathematical models: \textit{Mathematical Model} as the central class and anchor, \textit{Application Domain} as the domain in which the model is used, \textit{Application Problem} as the application problem that shall be solved by using the model, \textit{Mathematical Formulation} that formalizes the model (e.g., a \LaTeX~formula) as well as the \textit{Quantities} that enter into the mathematical formulation. This ontology for mathematical models, named MathModDB, consists of these five main classes and is shown in dark blue in fig.~\ref{fig:Ontology}.

Referring back to the X-RCT use case, the MathModDB ontology can be used to represent that the \textit{Mathematical Model} of X-ray transformation  \textit{models} the \textit{Application Problem} of microfracture detection of porous media, which in the specific case of asphalt \textit{is contained in} the \textit{Application Domain} of civil engineering. 
Moreover, it \textit{contains} a \textit{Mathematical Formulation}, which is  \textit{generalized by} the transport equation, and contains \textit{Quanities} such as radiant energy.

\paragraph{\textbf{Algorithms}} are the basic building block of applied mathematics. They are at the center of the AlgoData ontology \cite{AlgoData2022}, which is visualized in light blue in fig.~\ref{fig:Ontology}.

For instance, starting from the mathematical problem given above, inversion of the X-ray transform, the KG then provides a structured path to the relevant algorithms (e.g. filtered back projection and discrete algorithms), to scientific papers that have \textit{invented}, \textit{studied} or \textit{analyzed} them, to software packages \textit{implementing} them, and benchmarks \textit{testing} them.
Algorithms are grouped in classes based on common properties or algorithmic ideas, which makes the analysis easier and reveals connections between algorithms that were challenging to observe before.

\par
As with X-RCT, user stories for the AlgoData graph start with an application-specific problem formulation.
The \textit{Algorithmic Problems} considered in AlgoData are specified with respect to their mathematical properties, stating assumptions made on involved matrices, vector spaces, expected output etc.
The level of detail for these specifications varies between mathematical disciplines. For instance, in the context of model order reduction, very fine-grained specification is typically neither possible nor desired.
Hence, different fields of applied mathematics require different notions of problems to be able to incorporate their algorithms.
AlgoData does not cover the application specific part of the problem. To become useful, it must be connected to an ontology that defines application problems and corresponding models, i.e. MathModDB.

\paragraph{\textbf{Connecting the MathModDB and AlgoData Ontologies}}
In order to later exploit potential synergies of the two (and other) ontologies, we connect the MathModDB and the AlgoData ontologies.
The merging was done after the parallel development of the single ontologies for algorithms and models and the result is shown in fig.~\ref{fig:Ontology}.
Solving a mathematical problem, that has been derived from an application problem via modeling, represented in MathModDB, requires solving algorithmic subproblems, represented in AlgoData.
This link is provided by the relation/object property \textit{uses algorithmic problem} and its inverse \textit{used by model problem}.
This relation represents the process from the MSO workflow in which the mathematical modeling step is followed by solving the equations formalizing a model using an algorithm (cf.\ fig.~\ref{fig:MSO_Workflow}), which is however mediated through the class of \textit{Algorithmic Problems}.
Making such a link for an application requires finding the most specific AlgoData algorithmic problem object, taking into account the problem properties like symmetry. 
\par
Linking the mathematical model to the algorithmic problem rather than directly to the algorithms is beneficial: The initial collection of available algorithms needs to be done only once for each problem class, and new algorithms for a specific algorithmic subproblem immediately become available for all applications that use it, without an explicit link to the mathematical formulation.

However, the distinction between a mathematical problem and an algorithmic subproblem is diffuse. \textit{Mathematical Models} in the MathModDB ontology can be so specific and have such peculiar properties, that no suitable corresponding \textit{Algorithmic Problems} in the AlgoData graph exist. In these cases the mathematical problem should be added to AlgoData as an \textit{Algorithmic Problem}. 
\par
The current state of this implementation of the KG based on the ontologies is available on the beta version of the AlgoData site~\cite{AlgoData}. The interface includes documentation of the ontology, keyword search, and a guided query tailored to the ontology. Technically, the frontend is based on an \textit{Apache Jena Fuseki} with OWL reasoner and SPARQL interface, operated on a \textit{Django server}.
\par

The GUI invites the user to browse the data, discovering new relations and alternatives. Returning to the X-RCT working example from sec.~\ref{sec:Introduction}, we can now include a complete MSO workflow in the KG and show the full workflow chain starting in the MathModDB from the \textit{Application Domain} (civil engineering), choosing the \textit{Application problem} (porous media analysis/micro fracture detection), choosing the mathematical model of the transport equation and its specialization X-ray transform, moving over to the AlgoData application problem of its inversion, retrieving the basic \textit{Algorithms} classes (filtered back projection, algebraic reconstruction technique) and their implementations.

\section{Conclusion and Outlook}
\label{sec:Conclusion}
In our work, we show the development of the two ontologies for mathematical models and algorithms. Since both artifacts have a central place in the MSO workflow, but cannot be separated from each other for a holistic knowledge representation, they are merged. While the ontology for algorithms already contained a lot of data, since its development started earlier, for the merging we used the X-RCT use case to demonstrate what the KG can accomplish, namely representing knowledge of a concrete MSO workflow. The ontologies presented here allow a standardized description of mathematical models and algorithms. We will contribute to the \textit{WikiProject Mathematics} to further improve their representation in \textit{Wikidata}.

As future work, we intend to link our ontology to existing approaches. It is planned that \textit{Quantities, Units, Dimensions and Types} (QUDT)~\cite{QUDT2023}  as an elaborated data model and a vocabulary will be used for the class of \textit{Quantities}. For the \textit{Application Domain}, subject classification ontologies are currently being analyzed, such as the  \textit{Mathematics Subject Classification} (MSC) or the \textit{Physics Subject Headings} (PhysSH). The \textit{Publication} class is evaluated to be outsourced to a scholarly KG~\cite{Limani2021,Verma2023}. A class like \textit{Application Problem} can later be used to link to other discipline-specific ontologies from the NFDI. 
 
 Furthermore, to represent the steps of the MSO workflow, the embedding in a process ontology is planned. 
 Here, the m4i ontology is highly relevant as it offers a general process model that reuses specific instances for methods and tools from other, discipline-specific terminologies. The presented work on mathematical models, algorithms, implementations and benchmarks are ideal entry points for that. Furthermore, the embedding in upper ontologies (e.g. BFO~\cite{Otte2022}), as well as the inclusion of epistemic metadata (e.g. PIMS-II~\cite{Horsch2022}) is envisioned.  
After clarification of these open points, the ontology will be continuously instantiated with further data, becoming a living knowledge graph and serving as a knowledge base for mathematicians, application domain experts and beyond.

\subsection*{Acknowledgements}
The co-authors B.S., C.B., J.F., M.R., A.S., B.Sch. acknowledge funding by MaRDI, funded by the DFG (German Research Foundation), project number 460135501, NFDI 29/1 “MaRDI – Mathematische Forschungsdateninitiative”. The co-authors F.W. and H.K. acknowledge funding by the DFG under Germany's Excellence Strategy EXC 2044-390685587, Mathematics Münster: Dynamics–Geometry–Structure.
The co-author D.G. acknowledges funding by the DFG under Germany's Excellence Strategy EXC 2075:  Data-Integrated Simulation Science (SimTech), project number 390740016.

 \bibliographystyle{splncs04}
  \bibliography{MTSR_Preprint}

\end{document}